\documentclass[letterpaper]{article} 
\usepackage{aaai24}  
\usepackage{times}  
\usepackage{helvet}  
\usepackage{courier}  
\usepackage[hyphens]{url}  
\usepackage{graphicx} 
\usepackage{natbib}  
\usepackage{caption} 
\usepackage{algorithm}
\usepackage{algorithmic}

\usepackage{newfloat}
\usepackage{listings}
\DeclareCaptionStyle{ruled}{labelfont=normalfont,labelsep=colon,strut=off} 
\floatstyle{ruled}
\newfloat{listing}{tb}{lst}{}
\floatname{listing}{Listing}

\usepackage{amsmath}
\usepackage{amsfonts}
\usepackage{booktabs}
\usepackage{multirow}

\title{Medical Dialogue Generation via Intuitive-then-Analytical Differential Diagnosis}

\author{
    Kaishuai Xu\textsuperscript{\rm 1}, Wenjun Hou\textsuperscript{\rm 1,2}, Yi Cheng\textsuperscript{\rm 1}, Jian Wang\textsuperscript{\rm 1}, Wenjie Li\textsuperscript{\rm 1}
}
\affiliations{
    \textsuperscript{\rm 1}Department of Computing, The Hong Kong Polytechnic University, Hong Kong\\
    \textsuperscript{\rm 2}Research Institute of Trustworthy Autonomous Systems and\\
    Department of Computer Science and Engineering,\\
    Southern University of Science and Technology, Shenzhen, China\\

    \{kaishuaii.xu, alyssa.cheng, jian-dylan.wang\}@connect.polyu.hk, \\
     houwenjun060@gmail.com, cswjli@comp.polyu.edu.hk
}

\usepackage{bibentry}

\begin{document}

\maketitle

\begin{abstract}

Medical dialogue systems have attracted growing research attention as they have the potential to provide rapid diagnoses, treatment plans, and health consultations. In medical dialogues, a proper diagnosis is crucial as it establishes the foundation for future consultations. Clinicians typically employ both intuitive and analytic reasoning to formulate a differential diagnosis. This reasoning process hypothesizes and verifies a variety of possible diseases and strives to generate a comprehensive and rigorous diagnosis. However, recent studies on medical dialogue generation have overlooked the significance of modeling a differential diagnosis, which hinders the practical application of these systems. 
To address the above issue, we propose a medical dialogue generation framework with the Intuitive-then-Analytic Differential Diagnosis (IADDx). Our method starts with a differential diagnosis via retrieval-based intuitive association and subsequently refines it through a graph-enhanced analytic procedure. The resulting differential diagnosis is then used to retrieve medical knowledge and guide response generation. Experimental results on two datasets validate the efficacy of our method. Besides, we demonstrate how our framework assists both clinicians and patients in understanding the diagnostic process, for instance, by producing intermediate results and graph-based diagnosis paths. 

\end{abstract}
 
\section{Introduction}

\begin{figure}[th!]
	\centering
	\includegraphics[width=0.93\linewidth]{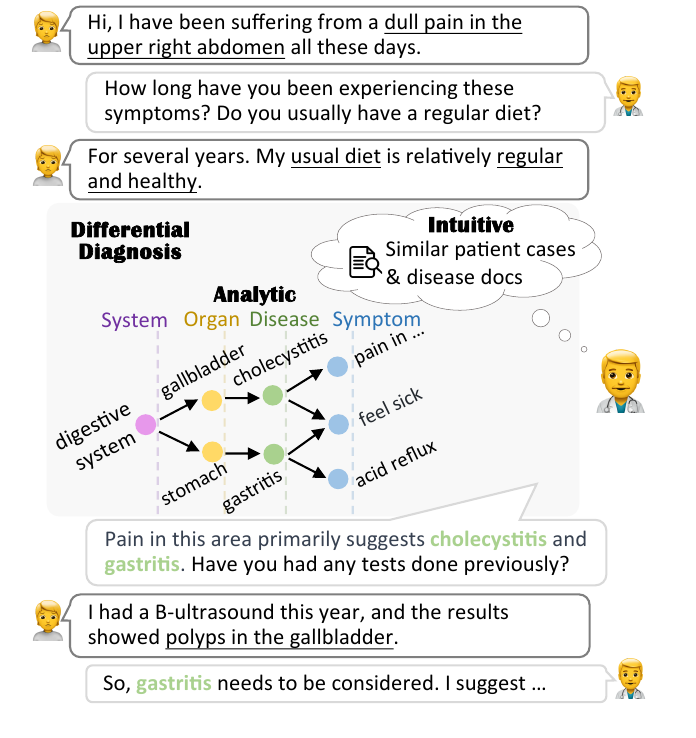}
	\caption{An example of differential diagnosis in a medical dialogue, which contains intuitive and analytic reasoning.}
	\label{example}
\end{figure}

Medical dialogue systems (MDS) endeavor to provide diverse medical services such as diagnosis, treatment plans, and health consultations \cite{remedi, dfmed, benchmark_mds}. These systems have garnered increasing research attention in recent years due to their potential to assist clinicians in diagnosing and prescribing \cite{midmed, meddg, meddialog, covid-19-meddialog, med_recommend, kg-routed-diagnosis}. 

In medical dialogues, diagnosis is a crucial process as the results of diagnosis establish an essential foundation for subsequent consultations \cite{analysis_medical_com, skills_for_medical_com}. However, previous studies on medical dialogue generation using pre-trained language models neglected to explicitly model the diagnostic process \cite{dfmed, med_pivotal, vrbot, hetero, geml, meddg}. One significant issue with these methods is that although responses benefiting from pre-trained models may appear coherent, they usually lack an interpretation grounded in meticulous medical diagnosis. It is challenging for clinicians or patients to accept responses from MDSs without a clear and interpretable diagnostic basis \cite{teach_med}. 

In practice, clinicians typically employ both intuitive and analytic reasoning during the dialogue to formulate a differential diagnosis, i.e., a set of potential diseases guiding how the subsequent dialogue unfolds \cite{diagnostic_model, trust_med, skills_for_medical_com}. Intuitive reasoning forms a rough disease list through a quick review of extensive clinical experience, while analytic reasoning cautiously verifies some diseases via a systematic analysis of body systems, organs, and symptoms. 
As an example shown in Figure \ref{example}, if a patient is diagnosed with a high possibility of gastritis but may still have chronic cholecystitis, the physician will first ask if any tests have been done to rule out chronic cholecystitis and then inquire about more gastritis-related symptoms to prescribe medications. 
Prior studies often overlook the importance of differential diagnosis \cite{vrbot, med_pivotal, kg-routed-diagnosis, tod-diagnosis, symp_check}. In our work, we argue that modeling a differential diagnosis with intuitive and analytic reasoning is crucial, and generating responses conditioned on potential diseases can improves the reliability and accuracy of medical dialogue generation. 

To address the above issues, we propose a medical dialogue generation framework with \textbf{I}ntuitive-then-\textbf{A}nalytic \textbf{D}ifferential \textbf{D}iagnosis (IADDx), which first produces a differential diagnosis and then utilizes potential diseases to guide response generation. 
For differential diagnosis, we draw inspiration from the diagnostic reasoning research \cite{diagnostic_model} and design a two-stage (i.e., intuitive-then-analytical) differential diagnosis method. In the intuitive stage, we extract patients' conditions from the dialogue and use them to retrieve previous cases and disease documents that present similar situations. A preliminary list of potential diseases can be concluded from the cases and documents. In the analytical stage, we first create a diagnosis-oriented entity graph that contains body systems, organs, diseases, and symptoms.
Then, we employ ConceptTransformer \cite{concept_trans} to incorporate the constructed graph and build a multi-disease classifier to discriminate multiple diseases, thereby assisting in refining the preliminary list. Our analytical stage achieves a multi-disease classification and provides a faithful and plausible interpretation represented by entities on the graph. 
For response generation, we utilize refined potential diseases to retrieve medical knowledge and generate responses conditioned on the knowledge. 

Our main contributions are summarized as follows:
\begin{itemize}

\item We propose a medical dialogue generation framework, IADDx, which explicitly models a differential diagnosis with intuitive-then-analytic reasoning and incorporates diagnosis to guide response generation. 

\item We build a diagnosis-oriented entity graph composed of systems, organs, diseases, and symptoms and apply the graph to enhance and interpret the diagnostic process in conversations. 

\item Experimental results on two medical datasets show the effectiveness and interpretability of our IADDx. 

\end{itemize}

\section{Related Work}

Medical dialogue systems (MDS) strive to offer healthcare services to patients. Initial research concentrated on automated diagnosis through task-oriented dialogue systems, emphasizing the swift identification of latent symptoms and providing a final diagnosis \cite{tod-diagnosis-hrl,symp-graph,diaformer,br_agent}. The work of \citet{tod-diagnosis} introduced a dataset marked with symptom annotations and developed a medical dialogue system using reinforcement learning. \citet{kg-routed-diagnosis} integrated a medical knowledge graph into MDS to manage the order of inquired symptoms. \citet{trust_med} further improves system reliability by outputting a differential diagnosis, using the exploration-conﬁrmation method, and prioritizing serious diseases. However, these systems conclude with diagnostic results without providing treatment plans or consultations. 

The advent of large-scale medical dialogue datasets like MedDialog \cite{meddialog}, MedDG \cite{meddg}, and KaMed \cite{vrbot} and pre-trained language models \cite{bart,gpt-2} have amplified interest in medical dialogue generation with multiple services. The study by \citet{meddg} approached medical dialogue generation by focusing on entity prediction coupled with entity-centric response generation. Moreover, \citet{hetero} enhanced dialogue understanding and entity reasoning using a unified heterogeneous graph. Similarly, \citet{med_pivotal} construct a dialogue graph to leverage medical relationships implied in the context. \citet{vrbot} treated medical entities within both patient and doctor utterances as states and actions, introducing a semi-supervised variation reasoning system complemented by a patient state tracker and a physician action network. \citet{dfmed} proposed a dual flow (i.e., dialogue act and entity flows) modeling method to improve dialogue understanding and use acts and entities to guide response generation. \citet{geml} explored to transfer the diagnostic experience from rich-resource diseases to low-resource ones. 

Although previous studies on medical dialogue generation have attempted to enhance dialogue understanding and guide response generation by incorporating predicted dialogue acts and medical entities, they ignore modeling the diagnostic process, which provides interpretation for generated responses. Besides, few works focus on differential diagnosis and apply it to instruct response generation. Our framework aims to model the process of differential diagnosis and generate responses with diagnosis hints. 

\section{Preliminary}

\begin{figure*}[th!]
	\centering
	\includegraphics[width=0.84\linewidth]{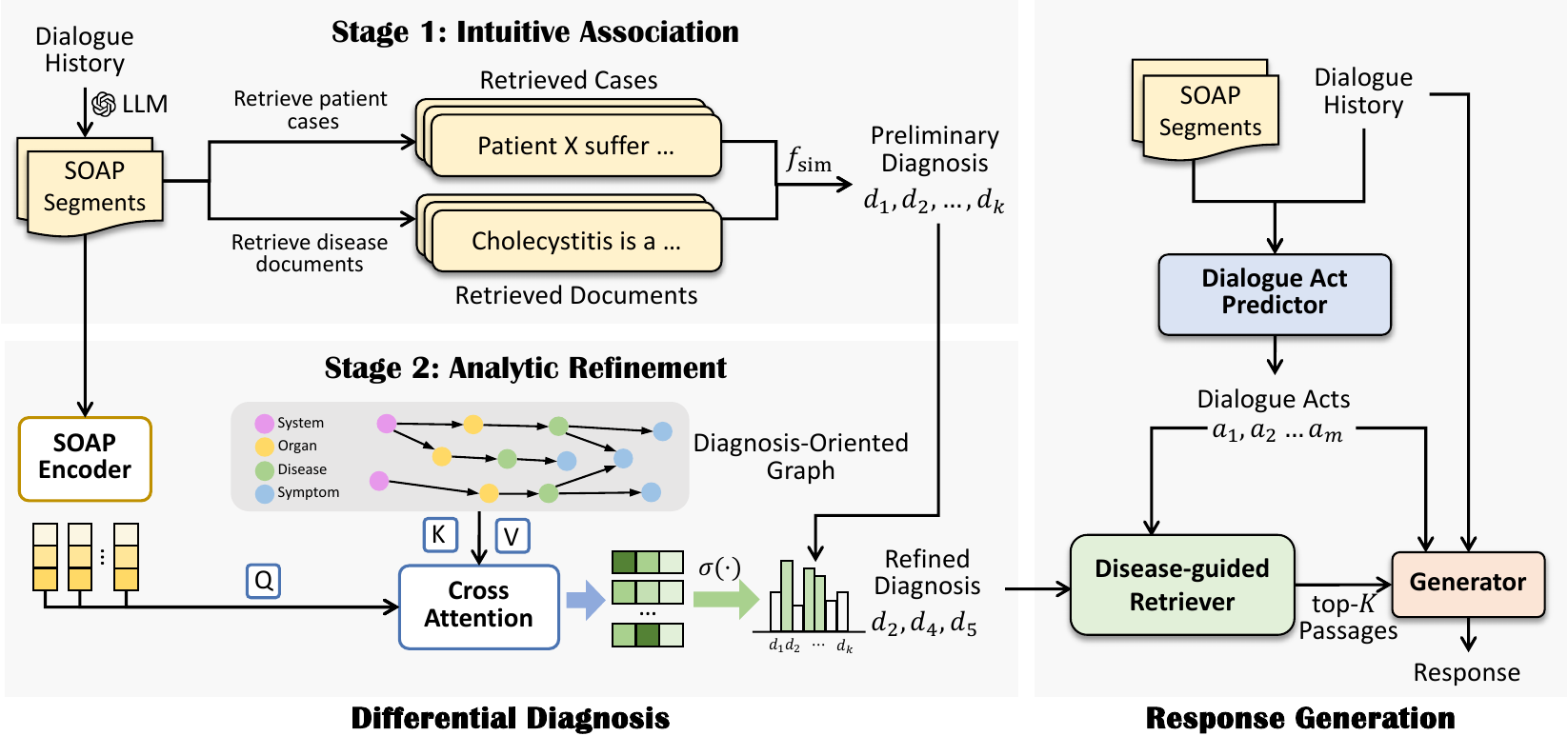}
	\caption{\textbf{Left}: The architecture of Differential Diagnosis, which includes the intuitive association stage and the analytic refinement stage. The multi-disease classifier in Stage 2 generates a refined diagnosis to guide response generation. \textbf{Right}: The structure of Response Generation. The diagnosis combined with the dialogue acts are used to retrieve relevant knowledge.}
	\label{framework}
\end{figure*}

\subsection{Problem Formulation}

A medical dialogue is denoted as $U$$=$$\{(U^P_k, U^D_k)\}_{k=1}^T$, where utterances from patients and doctors are represented by $U^P$ and $U^D$ respectively. Each dialogue is annotated with several possible diseases $D$$=$$\{d_i\}_{i=1}^{n_d}$, and each doctor utterance is annotated with multiple dialogue acts $A$$=$$\{a_i\}_{i=1}^{n_a}$. Given the historical dialogue sequence $U_t$$=$$\{U^P_1, U^D_1, ..., U^P_t\}$, the objective is to produce the $t$-th doctor utterance $U^D_t$. 

\subsection{SOAP-based Content Structuring}

In each round of a medical dialogue $(U^D, U^P)$, we first extract medical-related segments in accordance with the SOAP Notes \cite{soap_summar}, i.e., \textbf{S}ubjective personal reports $\{S^S_i\}$, \textbf{O}bjective quantifiable data $\{S^O_i\}$, an \textbf{A}ssessment (or diagnosis results) $\{S^A_i\}$, and subsequent \textbf{P}lans $\{S^P_i\}$ for patient care. 
SOAP notes serve as a documentation method used by medical specialists to structure patient information. 
For example, a subjective segment can be ``vomited three times''. The extraction filters out irrelevant details, and segments from the subjective and objective sections are instrumental for diagnosis. We employ pre-trained large language models (LLMs), such as GPT-4 \cite{gpt_4}, to extract segments in a few-shot manner. The input prompts for LLMs contain SOAP instructions and some examples of extracted segments. Details are described in Appendix.

\subsection{Diagnosis-Oriented Graph Construction}

We construct a Diagnosis-Oriented Graph (DOG) $G$$=\{ e_i \}$ inspired by the \textit{problem-specific framework} used in differential diagnosis \cite{symptom_to_diagnosis}. This framework aids in pinpointing medical issues and inferring a differential diagnosis. The graph we develop encompasses entities such as body systems ($e^{\text{Sys}}$), organs ($e^{\text{Org}}$), diseases ($e^{\text{Dis}}$), and symptoms ($e^{\text{Sym}}$). Notably, ``System $\xrightarrow{}$ Organ $\xrightarrow{}$ Disease $\xrightarrow{}$ Symptom'' can be a diagnostic path. In detail, systems and organs serve to categorize disease types and help understand the potential impact, while symptoms provide the basis for confirming or ruling out a particular disease. Based on findings from anatomic studies \cite{anatomy}, we incorporate major body systems, such as the digestive and endocrine systems, and associated organs like the stomach and thyroid. Besides, we select diseases and symptoms from an online medical encyclopedia website\footnote{https://www.baikemy.com/} that is edited and reviewed by medical specialists. The relations between body systems and their respective organs, as well as between diseases and symptoms, are inherent. To establish connections between diseases and the organs they affect, we utilize pre-trained LLMs to associate diseases with specific organs based on their pathological manifestations. 

\section{Method}

Our proposed IADDx framework comprises two main components, as illustrated in Figure \ref{framework}. The Intuitive-then-Analytic Differential Diagnosis component makes a differential diagnosis through intuitive and analytic reasoning. 
Subsequently, the Diagnosis-guided Response Generation component utilizes diagnosis results to guide knowledge retrieval and generate an appropriate response with the retrieved knowledge. 

\subsection{Ituitive-then-Analytic Differential Diagnosis}

We model the differential diagnosis process in two stages: \textbf{intuitive association} and \textbf{analytic refinement}. 
The first stage draws upon clinical experience to make a rough diagnosis. We use patient case and disease document retrieval to generate a preliminary list of diseases. The disease document retriever assists in identifying diseases that have not been previously encountered. 
Then, the second stage further refines the diagnosis through a more detailed problem analysis, enhancing both diagnostic accuracy and interpretability. We construct a multi-disease classifier with the aid of our \textbf{diagnosis-oriented graph} for this stage. 

\subsubsection{Stage 1: Intuitive Association.}
The objective of this stage is to make a rough diagnosis. 
Given the dialogue history $U_t$, we extract SOAP segments $\{S^S_i\}$, $\{S^O_i\}$, $\{S^A_i\}$, and $\{S^P_i\}$ using a pre-trained LLM and use them to retrieve knowledge for disease list generation. These segments contain information (i.e., patient symptoms, signs, and medical history) needed for differential diagnosis. We concatenate the $\{S^S_i\}$, $\{S^O_i\}$, and $\{S^A_i\}$ segments as the query for subsequent retrieval. Here, the plan segments are not considered since their content is presented after diagnosis. We use the aforementioned query to retrieve two types of knowledge: (1) patient cases (in the format of SOAP) from the training corpus exhibiting conditions similar to those of the current patient and (2) disease documents (i.e., etiology and symptoms) from a disease corpus with descriptions that align with the current patient's conditions. 
The query and patient case (or disease document) sequences with a ``[CLS]'' token inserted at the front are separately input to a BERT encoder \cite{bert}. We select hidden states of ``[CLS]'' tokens for each sequence as their representations $\textbf{S} \in \mathbb{R}^d$ and $\textbf{S}^{\text{Case}}$$\in \mathbb{R}^d$ (or $\textbf{S}^{\text{Doc}}$$\in \mathbb{R}^d$) and calculate relevance scores as:
\begin{align}
s^{\text{Case}} = \langle \textbf{S}, \textbf{S}^{\text{Case}} \rangle, s^{\text{Doc}} = \langle \textbf{S}, \textbf{S}^{\text{Doc}} \rangle,
\end{align}
where $\langle, \rangle$ represents a similarity function. After retrieving patient cases, each case is assigned a relevance score, which is then applied to all diseases associated with that case. Thus, each disease can be assigned a group of scores $\{s^{\text{Case}}_j\}_{j \in {\cal{C}}_i}$ from different patient cases, where ${\cal{C}}_i$ denotes cases diagnosed with disease $i$. We select the maximum as the relevance score for each disease $s^{\text{Case}*}_i$. After retrieving disease documents, each disease can also be directly assigned a relevance score $s^{\text{Doc}}_i$. We average these two scores as the final disease relevance score, $s_i$$ = $$(s^{\text{Case}*}_i + s^{\text{Doc}}_i) / 2$, and select top-$K$ diseases to generate a preliminary disease list $\{ d_i \}_{i=1}^K$. 

We adopt Contrastive Learning \cite{simclr} to optimize these two retrievers. The loss function for the patient case retriever is defined as: 
\begin{align}
{\cal{L}}_{\text{Case}} &= - \log \frac {\exp ( \langle \textbf{S}, \textbf{S}^{\text{Case}+}_t \rangle ) }{ \sum_{\textbf{S}^{\text{Case}-}_t \in \cal{B}} \exp ( \langle \textbf{S}, \textbf{S}^{\text{Case}-}_t \rangle ) } ,
\end{align}
where $\textbf{S}^{\text{Case}+}_t$ denotes representations from positive cases that share at least one diagnosed disease with the current dialogue, and $\textbf{S}^{\text{Case}-}_t$ denotes those from negative cases that do not coincide with diseases discussed in the current dialogue. We use negative cases from input batches $\cal{B}$ for training. The loss function ${\cal{L}}_{\text{Doc}}$ for the disease document retriever is defined in the same way.

\subsubsection{Stage 2: Analytic Refinement.}
This stage further refines the diagnosis via specific problem analysis, improving diagnostic accuracy and interpretability. We construct a multi-disease classifier inspired by the ConceptTransformer \cite{concept_trans} model for this stage, which leverages domain knowledge to improve classification accuracy and provide concept-based interpretations. The diagnosis-oriented graph is employed to augment and interpret the classification.

As depicted in the left of Figure \ref{framework}, the inputs of the classifier are SOAP segments extracted in the first stage and the diagnosis-oriented graph. Specifically, SOAP segments are encoded into several representations $\{ \textbf{S}_i \}_{i=1}^{n_s}$ through a BERT encoder, and each one is obtained by averaging hidden states of tokens corresponding to one segment. $n_s$ is the number of no duplicate segments until the current turn. 
We incorporate entities that are involved in the diagnostic path of diseases identified in the preliminary diagnosis and obtain the sub-graph $G_t$. For all entities in the sub-graph, we use the same encoder as SOAP segments to get entity embeddings. The average token embedding of each entity is utilized as the raw embedding, denoted as $\textbf{e}^0 \in \mathbb{R}^d$. We employ Graph Attention Network (GAT) \cite{gat} to merge neighboring information for each entity:
\begin{align}
 \alpha^k_{ij} &= \frac{\exp \left(\sigma_1 \left(\textbf{a}^{\mathsf{T}} [\textbf{W}^k \textbf{e}_i^0 \Vert \textbf{W}^k \textbf{e}_j^0 ] \right) \right)}
 {\sum_{\mu \in \mathcal{N}_i}\exp \left(\sigma_1 \left(\textbf{a}^{\mathsf{T}} [\textbf{W}^k \textbf{e}_i^0 \Vert \textbf{W}^k \textbf{e}_{\mu}^0 ] \right) \right)}, \\
 \textbf{e}_i &= \left[ \sigma_2 \left(\sum_{j \in \mathcal{N}_i} \alpha^k_{ij} \textbf{W}^k \textbf{e}_j^0 \right) \right]_{k=1}^h,
\end{align}
where $\textbf{e}_i \in \mathbb{R}^d$ represents the updated embedding, $\textbf{a} \in \mathbb{R}^{2d}$ and $\textbf{W}^k \in \mathbb{R}^{d_h \times d}$ are learnable parameters, $\sigma_1$ and $\sigma_2$ denote activation function, $\mathcal{N}_i$ is a set of neighboring entities that connect to entity $i$, and $h$ is the number of heads. 

The multi-disease classifier incorporates a cross-attention layer to facilitate interaction between segment representations and entity embeddings. Each segment representation integrates relevant entity information from the graph, and this enriched representation is then transformed via linear mapping to estimate probabilities for multiple diseases. 
The attention matrix $\textbf{A}$ and the probability $p^d_j$ for each disease are calculated as follows:
\begin{align}
\textbf{A} &= \text{softmax}(\frac{{\textbf{Q}} {\textbf{K}}^{\mathsf{T}}}{\sqrt{d}}), \\
p^d_j &= \text{sigmoid} ( \sum ^{n_s}_{i=1} [\textbf{A}{\textbf{V}}\textbf{O}]_{ij} ), j=1,\dots,n,
\end{align}
where $\textbf{K} \in \mathbb{R}^{n_{G_t} \times d}$ and $\textbf{V} \in \mathbb{R}^{n_{G_t} \times d}$ are the linear projected matrix based on the concatenation of entity embeddings $\{\textbf{e}_i \}_{i \in G_t}$, and $\textbf{Q} \in \mathbb{R}^{n_s \times d}$ is based on the segment representations $\{ \textbf{S}_i \}_{i=1}^{n_s}$. $\textbf{O} \in \mathbb{R}^{d \times n}$ denotes an output projection matrix, and $n$ is the number of total diseases in our corpus. 
We select the probabilities of diseases within the preliminary list and employ a proper threshold to predict multiple diseases (See in Experiments) as the refined differential diagnosis $\{ {d'}_i\}^{n_d}_{i=1}$. 

We optimize the multi-disease prediction by minimizing a binary cross-entropy loss. The loss function ${\cal{L}}_d$ is calculated as follows:
\begin{align}
{\cal{L}}_d = - \frac{1}{K} \sum [y^d_j \cdot \log p_j^d + (1 - y^d_j) \cdot \log (1 - p_j^d)],
\end{align}
where $y^d_j$ is the label of $j$-th disease, and $K$ denotes the number of diseases in the preliminary list. We optimize the disease probabilities within each list, which vary across dialogues. Besides, we add an explanation loss \cite{concept_trans} to supervise the attention weights. This loss can guide the attention heads to attend to entities that are beneficial for disease classification. The loss ${\cal{L}}_{expl}$ is defined as:
\begin{align}
{\cal{L}}_{expl} = \Vert \textbf{A} - \textbf{A}' \Vert_F^2,
\end{align}
where $\Vert \cdot \Vert_F$ denotes the Frobenius norm, $\textbf{A}'$ is the desired distribution of attention. The final loss used for training multi-disease classification is defined as follows:
\begin{align}
{\cal{L}} = \alpha {\cal{L}}_d + \beta {\cal{L}}_{expl}, 
\end{align}
where $\alpha$ and $\beta$ are weights for balancing these two losses.

\subsection{Diagnosis-guided Response Generation}

In this section, we utilize the refined diagnosis to guide medical knowledge retrieval, thereby enhancing response generation with the aid of retrieved disease-related knowledge. The diagnostic results help to select knowledge more accurately and guide the dialogue around related diseases. 
A doctor dialogue act predictor is introduced since acts help to select a specific aspect of disease knowledge (e.g., clinical manifestations or examinations) and guide the flow of the dialogue \cite{dfmed}. The detailed architecture is shown on the right of Figure \ref{framework}. 

\subsubsection{Dialogue Act Prediction.}
We develop a multi-act predictor to assist knowledge retrieval and manage the dialogue. The input of the predictor is SOAP segments and dialogue history. We concatenate ``[CLS]'' token at the front of the segment sequence and encode this sequence using a BERT encoder, which is also applied to the dialogue history. The final hidden states of ``[CLS]'' token in these two encodings are selected as the representation of the segments $\textbf{H}_t^s \in \mathbb{R}^d$ and dialogue history $\textbf{H}_t \in \mathbb{R}^d$. We merge the structured patient information and dialogue content to predict dialogue acts. The act probability is calculated as follows:
\begin{align}
p^a_i &= \text{sigmoid} ( \textbf{W}^a [\textbf{H}_t^s; \textbf{H}_t]), i=1,\dots,m,
\end{align}
where $\textbf{W}^a \in \mathbb{R}^{m \times 2d}$ is a trainable parameter matrix, $[;]$ denotes a concatenation operation, and $m$ is the number of candidate acts. The predicted dialogue acts $\{ a_i \}_{i=1}^{n_a}$ are selected through an appropriate threshold (See in Experiments). 

We apply a binary cross-entropy loss to optimize the multi-act prediction. The loss function is denoted as follows:
\begin{align}
{\cal{L}}_a = - \frac{1}{m} \sum [y^a_i \cdot \log p_i^a + (1 - y^a_i) \cdot \log (1 - p_i^a)],
\end{align}
where $y^a_i$ is the label of $i$-th dialogue act. 

\subsubsection{Disease-guided Retrieval.}

To augment response generation, we retrieve disease-related passages based on the refined differential diagnosis and predicted dialogue acts. These passages are from the medical encyclopedia website, providing external knowledge for response generation. 
For dialogue acts directly related to a specific aspect of medicine, we choose corresponding passages without requiring retrieval. For example, responses with the act ``Inquire about present illness'' are closely related to the passage describing clinical manifestations of one disease. For non-medical dialogue acts, we use the current dialogue history to retrieve relevant passages from a disease corpus. We remove stop words and punctuation from the dialogue history sequence and retrieve top-$k$ passages through the BM25 algorithm \cite{bm25}. 

\subsubsection{Response Generation.}
After retrieving medical knowledge and predicting dialogue acts, we incorporate these two pieces of information to guide response generation. We construct a generation model following the Fusion-in-Decoder (FiD) method \cite{fid, fid_conv}, which allows the decoder to attend to all encoding representations at the same time when generating a response. The input of the model contains a group of sequences, where the first sequence is the dialogue history sequence $U_t$ concatenated with unique tokens denoting dialogue acts $\{ a_i \}$, and other sequences are retrieved knowledge sequences $Z_t$. Each sequence is concatenated with a special token (``[U]'' or ``[K]'') at the front to represent the dialogue history or knowledge. Compared with concatenating the dialogue history with knowledge as an input sequence, the above input can leverage longer dialogue history. 

We train the generation model by a negative log-likelihood loss. The loss function is defined as:
\begin{align}
{\cal{L}}_g = - \sum_{i = 1}^N\log p(U^D_{t, i}),
\end{align}
where $U^D_{t, i}$ is the $i$-th token's probability. During training, we use ground truth acts as part of the input and apply ground truth diseases to guide knowledge retrieval. Then in inference, we apply predicted acts and diseases as guidance.

\section{Experiments}

\begin{table*}[th!]
\centering
\resizebox{0.74\linewidth}{!}{
\begin{tabular}{llcccccccc} 
\toprule
\multicolumn{2}{c}{\textbf{Methods}}                                   & \multicolumn{1}{l}{\textbf{B-1}} & \textbf{B-2} & \textbf{B-4} & \textbf{R-1} & \textbf{R-2} & \textbf{E-P} & \textbf{E-R} & \textbf{E-F1}  \\ 
\midrule
\multicolumn{1}{c}{\multirow{5}{*}{w/o Pre-training}} & Seq2Seq        & 28.55                            & 22.85        & 15.45        & 25.61        & 11.24        & 16.79        & 10.44        & 12.88          \\
\multicolumn{1}{c}{}                                  & Seq2Seq-Entity & 29.13                            & 23.22        & 15.66        & 25.79        & 11.42        & \textbf{23.79}        & 15.89        & 19.06          \\
\multicolumn{1}{c}{}                                  & HRED           & 31.61                            & 25.22        & 17.05        & 24.17        & 9.79         & 15.56        & 10.12        & 12.26          \\
\multicolumn{1}{c}{}                                  & HRED-Entity    & 32.84                            & 26.12        & 17.63        & 24.26        & 9.76         & 21.75        & 15.33        & 17.98          \\
\multicolumn{1}{c}{}                                  & VRBot          & 29.69                            & 23.90        & 16.34        & 24.69        & 11.23        & 18.67        & 9.72         & 12.78          \\ 
\midrule
\multirow{6}{*}{w/ Pre-trained LM}                    & GPT-2          & 35.27                            & 28.19        & 19.16        & 28.74        & 13.61        & 18.29        & 14.45        & 16.14          \\
                                                      & GPT-2-Entity   & 34.56                            & 27.56        & 18.71        & 28.78        & 13.62        & 21.27        & 17.10        & 18.96          \\
                                                      & BART           & 34.94                            & 27.99        & 19.06        & 29.03        & \textbf{14.40}        & 19.97        & 14.29        & 16.66          \\
                                                      & BART-Entity    & 34.14                            & 27.19        & 18.42        & 28.52        & 13.67        & 23.49        & 16.90        & 19.66          \\
                                                      & DFMed          & 42.56                            & 33.34        & 22.53        & 29.31        & 14.21        & 22.48        & 22.84        & \textbf{22.66}          \\
                                                      & IADDx (Ours)   & \textbf{43.17}$^\dag$           & \textbf{34.09}$^\dag$ & \textbf{23.33}$^\dag$ & \textbf{29.60}$^\dag$ & 14.37 & 21.81 & \textbf{22.90}$^\dag$  & 22.34          \\
\bottomrule
\end{tabular}
}
\caption{Automatic evaluation results on MedDG. † denotes statistically significant differences ($p$ = 0.05).}
\label{meddg_results}
\end{table*}

\subsection{Datasets}

We perform our experiments using two medical dialogue datasets: \textbf{MedDG} \cite{meddg} and \textbf{KaMed} \cite{vrbot}. 
MedDG encompasses 17K dialogues, primarily centered on 12 diseases within the gastroenterology department. We partition the dataset into 14862, 1999, and 999 dialogues for training, validation, and testing, respectively. 
KaMed offers a comprehensive collection of over 63K dialogues spanning nearly 100 hospital departments. To address privacy concerns (See in Appendix), we exclude certain dialogues from KaMed, resulting in 29159, 1532, and 1539 dialogues for training, validation, and testing, respectively. 

\subsection{Baseline Models}

We evaluate IADDx with six baseline models. 
\textbf{Non-Pretrained models:}
(1) \textbf{Seq2Seq} \cite{seq2seq} employs an RNN for sequence-to-sequence generation enhanced by an attention layer. (2) \textbf{HRED} \cite{hred} leverages a multi-level RNN design to encode dialogues both at the token and utterance levels. (3) \textbf{VRBot} \cite{vrbot} is designed for medical dialogue generation, emphasizing the tracking and predicting of patient and doctor entities.
\textbf{Pretrained models:}
(1) \textbf{GPT-2} \cite{gpt-2} is a transformer decoder-based language model. (2) \textbf{BART} \cite{bart} is a transformer-based encoder-decoder model. (3) \textbf{DFMed} \cite{dfmed} is a medical dialogue generation model that learns entity and dialogue act flows. 
For our experiments on the MedDG dataset, we supplement models with entity hints as described by \citet{meddg}. This involves appending extracted medical entities to the end of the input sequence. 

\subsection{Evaluation Metrics}

\paragraph{Automatic Evaluation.}

BLEU \cite{bleu} and ROUGE \cite{rouge} scores across varied n-grams (specifically, \textbf{B-1}, \textbf{B-2}, \textbf{B-4}, \textbf{R-1}, and \textbf{R-2}) are utilized as metrics to evaluate the quality of generated responses. Additionally, in alignment with \citet{meddg}, we calculate the precision, recall, and F1 of entities mentioned in the responses, denoted as \textbf{E-P}, \textbf{E-R}, and \textbf{E-F1} respectively. We evaluate the accuracy of the differential diagnosis using the disease F1 score, denoted as \textbf{D-F1}. 

\paragraph{Human Evaluation.}

\begin{table}[t!]
\centering
\resizebox{0.93\linewidth}{!}{
\begin{tabular}{lccccc} 
\toprule
\textbf{Methods} & \textbf{B-1}   & \textbf{B-2}   & \textbf{\textbf{B-4}} & \textbf{\textbf{R-1}} & \textbf{\textbf{R-2}}  \\ 
\midrule
Seq2Seq          & 23.52          & 18.56          & 12.13                 & 23.56                 & 8.67                   \\
HRED             & 26.75          & 21.08          & 13.91                 & 22.93                 & 7.80                   \\
VRBot            & 30.04          & 23.76          & 16.36                 & 18.71                 & 7.28                   \\
GPT-2            & 33.76          & 26.58          & 17.82                 & 26.80                 & 10.56                  \\
BART             & 33.62          & 26.43          & 17.64                 & 27.91                 & 11.43                  \\
DFMed            & 40.20          & 30.97          & 20.76                 & 28.28                 & 11.54                  \\
IADDx (Ours)     & \textbf{40.98}$^\dag$ & \textbf{31.69}$^\dag$ & \textbf{21.35}$^\dag$ & \textbf{28.31}$^\dag$ & \textbf{11.67}$^\dag$    \\
\bottomrule
\end{tabular}
}
\caption{Automatic evaluation results on KaMed. † denotes statistically significant differences ($p$ = 0.05).}
\label{kamed_results}
\end{table}

We randomly selected 100 cases and engaged three physicians for manual evaluation. The performance of our IADDx is compared with various baseline models. Drawing upon previous studies \cite{meddg,vrbot}, we assess the generated responses using three metrics: sentence fluency (FLU), knowledge accuracy (KC), and overall quality (EQ). Each metric is on a 5-point Likert scale, ranging from 1 (poorest) to 5 (excellent).

\subsection{Implementation Details}

We apply the MedBERT\footnote{https://github.com/trueto/medbert}, a BERT-base model pre-trained on Chinese medical documents as the backbone for encoders in the differential diagnosis component. The disease corpus used for the intuitive association is extracted from an online medical encyclopedia website named \textit{baikemy}\footnote{https://www.baikemy.com/}, which contains medical knowledge certified by specialists. 
In the intuitive association stage, we selected the top 50 diseases as the preliminary diagnosis results. In the analytic refinement stage, we predict multiple diseases from the list. We set the weights for ${\cal{L}}_d$ and ${\cal{L}}_{expl}$ to 1 and 0.5. The threshold for predicting multiple diseases is set to 0.8 based on the disease F1 scores on the validation dataset. In the response generation component, we retrieve knowledge from the same corpus and select the top 5 passages for generating responses. The dialogue act predictor is also based on the MedBERT. We choose from 10 acts and adopt different thresholds to get the highest act F1 scores for each act. The generator is a pre-trained BART-base model\footnote{https://huggingface.co/fnlp/bart-base-chinese} with a six-layer encoder and a six-layer decoder. We adopt the AdamW optimizer \cite{adamw} to train the above models and implement all experiments on a single RTX 3090 GPU. Further training details are provided in Appendix.

\subsection{Evaluation of Dialogue Generation}

\begin{table}[t!]
\begin{center}
\resizebox{0.55\linewidth}{!}{
\begin{tabular}{@{}lccc@{}}
\toprule
\multicolumn{1}{c}{\textbf{Methods}} & \textbf{FLU} & \textbf{KC} & \textbf{EQ} \\ \midrule
BART                     & 3.77      &  1.87    & 3.12    \\
BART-Entity              & 3.79      &  2.05    & 3.41    \\
DFMed                    & 3.91      &  2.26    & 3.59    \\ 
IADDx (Ours)             & \textbf{4.08}  &  \textbf{2.41} & \textbf{3.83}  \\ 
\bottomrule
\end{tabular}
}
\end{center}
\caption{\label{human_eval} Human evaluation results on MedDG.}
\end{table}

\subsubsection{Automatic Evaluation.}
The dialogue generation results for the MedDG dataset are presented in Table \ref{meddg_results}, and results for the KaMed dataset can be found in Table \ref{kamed_results}. We observe that our IADDx method outperforms baseline models on most evaluation metrics. In detail, on the MedDG dataset, IADDx surpasses the state-of-the-art method DFMed by 0.61, 0.75, and 0.8 in B-1, B-2, and B-4 metrics, as well as 0.29 and 0.16 in R-1 and R-2 metrics. Additionally, we achieve comparable entity accuracy even without employing entity flow learning like DFMed. The reason is that we use multiple differential diagnoses to retrieve pertinent knowledge and subsequently leverage them to enhance response generation. 
Similar advantages are evident in the experimental results on the KaMed dataset. IADDx outperforms DFMed by 0.78, 0.72, and 0.59 in B-1, B-2, and B-4 metrics. It indicates that modeling and incorporating differential diagnosis aid in generating coherent, informative, and accurate responses. 

\subsubsection{Human Evaluation.}
Table \ref{human_eval} shows the results of human evaluation on the MedDG dataset. Our method outperforms baselines in all metrics. This suggests that by explicitly modeling the differential diagnosis and using it to guide response generation, we can produce more informative and accurate responses. The Fleiss’ kappa \cite{fleiss_kappa} score is 0.49, indicating a moderate level of inter-annotator agreement. 

\subsection{Analysis of Intuitive-then-Analytic Differential Diagnosis}

\begin{table}[t!]
\centering
\resizebox{0.93\linewidth}{!}{
\begin{tabular}{llcccc} 
\toprule
\textbf{Datasets}      & \textbf{Methods} & \multicolumn{1}{l}{\textbf{D-F1}} & \textbf{B-1} & \textbf{\textbf{B-4}} & \textbf{\textbf{R-2}}  \\ 
\cmidrule{1-6}
\multirow{4}{*}{MedDG} & IADDx            & 43.50                              & 43.17        & 23.33                 & 14.37                  \\
                       & ~~~~w/o DDx      & -                                 & 42.21        & 22.08                 & 13.78                  \\
                       & ~~~~w/o Analytic & 37.01                                  & 43.02        & 22.98                 & 14.09                  \\
                       & ~~~~w/o DOG      & 42.61                             & 43.13        & 23.22                 & 14.23                  \\ 
\cmidrule{1-6}
\multirow{4}{*}{KaMed} & IADDx            & 50.23                             & 40.98        & 21.35                 & 11.67                  \\
                       & ~~~~w/o DDx      & -                                 & 39.03        & 19.54                 & 10.22                  \\
                       & ~~~~w/o Analytic & 40.26                                  & 40.74        & 21.16                 & 11.44                  \\
                       & ~~~~w/o DOG      & 49.52                                  & 40.86        & 21.22                 & 11.52                  \\
\bottomrule
\end{tabular}
}
\caption{Ablation results on two datasets.}
\label{ablation}
\end{table}

To delve deeper into the efficacy of our approach, we examine multiple variants of our IADDx method as follows:
(1) \textbf{w/o DDx}, which removes the entire differential diagnosis component and generates responses conditioned solely on the dialogue history and dialogue acts. 
(2) \textbf{w/o Analytic}, which removes the analytic refinement on the preliminary diagnosis and employs the top 5 ranked diseases to guide knowledge retrieval. 
(3) \textbf{w/o DOG}, which removes the diagnosis-oriented graph in the multi-disease classifier and solely adopts the mean representations of SOAP segments and a subsequent linear layer to classify diseases. 

Table \ref{ablation} displays the overall ablation results. We see a reduction in performance across all metrics with the ablation variants, emphasizing the critical role of each module in our proposed method. Among these variants, the response quality of \textit{w/o} DDx notably decreases due to the absence of diagnosis-related knowledge. Such knowledge is crucial for providing essential disease information that enhances the informativeness and accuracy of the response. Besides, the differential diagnosis performance of \textit{w/o} Analytic significantly drops compared with the complete DDx. It is because the retrieval-based intuitive association provides a fixed number of potential diseases and cannot ensure that all relevant diseases rank high, inadvertently introducing unrelated diseases. The results of \textit{w/o} DOG demonstrate that the diagnosis-oriented graph plays a role in the differential diagnosis and subsequent diagnosis-guided response generation. 

\subsection{Case Study}

\begin{table}[!th]
\centering
\small
\begin{tabular}{p{0.95\linewidth}}
\toprule
\textbf{Patient}: Hello, doctor! Over the past week, I've been experiencing occasional subtle pain in the center of my upper abdomen, and sometimes this pain radiates to my back. Last night, I felt a tightness in my chest and a heavy sensation in my throat, almost like acid reflux. What could be the issue? \\
\textbf{Doctor}: Hello, are you currently experiencing acid reflux or heartburn? Do you burp often?  \\
\textbf{Patient}: I haven't had noticeable acid reflux, but there's a slight burning sensation. I started burping since yesterday. After taking medicine today, the burping has become less frequent. Sometimes, there's also a subtle burning feeling in my back. \\
...     \\
\textbf{Patient}: I didn't sleep well last night since I felt discomfort in my upper abdomen, and I woke up several times.  \\ 
\midrule
\textbf{Gold Response}: It's likely that you have \underline{gastroesophageal} \underline{reflux disease (GERD)}. I'd recommend trying Mosapride. For now, avoid greasy, fried, and spicy foods. \\ 
\midrule
\textbf{BART}: I suspect it's \underline{gastroesophageal reflux disease (GERD)}. \\
\textbf{BART-Entity}: Yes, it seems to be caused by an excessive secretion of stomach acid.  \\
\textbf{DFMed}: I believe you might have \underline{weak gastrointestinal} \underline{motility} and \underline{poor digestive function}. I recommend taking Omeprazole, Domperidone, and Mosapride for treatment. Stick to a bland diet and avoid spicy, irritating, and fried foods. \\
\midrule
\textbf{Differential Diagnosis}: (1) \underline{reflux esophagitis}, (2) \underline{gastro} \underline{-esophageal reflux}, (3) \underline{bile reflux gastritis}, (4) \underline{duodeno} \underline{-gastric reflux} \\
\textbf{Attended Entities in DOG}: [digestive system] $\xrightarrow{}$ [stomach] $\xrightarrow{}$ [\underline{reflux esophagitis} / \underline{gastroesophageal reflux}] $\xrightarrow{}$ [upper abdominal pain / poststernal burning sensation / acid reflux] \\ 
\textbf{IADDx (Ours)}: I suspect you have \underline{gastroesophageal reflux} \underline{disease (GERD)}. I recommend taking Rabeprazole, Mosapr-ide, and Magnesium aluminum carbonate for treatment. Over the next few days, be mindful of your diet. Focus on easily digestible foods, and try to avoid acidic, spicy, or cold foods. \\
\bottomrule
\end{tabular}
\caption{\label{case_study} Case study. The responses are generated by different models, where diseases are underlined.}
\end{table}

We present a case study of medical dialogue generation by different models in Table \ref{case_study}. We observe that the response generated by our IADDx method offers more accurate diagnoses, along with treatment plans tailored to those diagnoses, compared to responses from the three baseline models. The prescribed medicine, Rabeprazole, Mosapride, and Magnesium aluminum carbonate, are suitable for treating the diagnosed disease GERD. This suggests that IADDx effectively performs a reasonable differential diagnosis and leverages it to enhance response generation. Besides, although four diseases are included in the differential diagnosis, IADDx can implicitly select the most likely disease to generate a response. 

\subsubsection{Interpretation of differential diagnosis.}
Our IADDx offers a comprehensive list of differential diagnoses that can explain the generated response. In this dialogue, IADDx identifies four potential diseases that exhibit similar symptoms. The first and second diseases are both related to GERD and have been utilized to generate the response. Beyond these two, physicians still need to consider the potential of the other diseases and seek to confirm or rule them out in subsequent conversations. IADDx also provides a diagnosis path composed of systems, organs, diseases, and symptoms to interpret the differential diagnosis. We select entities with high attention weights to build the diagnosis path. We observe that the patient's condition predominantly pertains to the digestive system and the stomach. Additionally, the attention to specific symptom entities lends further support to the potential diseases: reflux esophagitis and gastroesophageal reflux. 

\section{Conclusion}

In this work, we propose a medical dialogue generation framework with a differential diagnosis, IADDx, which explicitly models the process of differential diagnosis through intuitive association followed by analytic refinement. Moreover, we devise a diagnosis-oriented graph to interpret the differential diagnosis. The diagnosis results are utilized to guide medical knowledge retrieval and response generation. Experiments on two datasets demonstrate the efficacy of the proposed framework. Additionally, we illustrate how our framework aids clinicians and patients in understanding the diagnostic procedure, such as by generating intermediate outcomes and graph-based diagnostic paths. 

\bibliography{aaai24}

\end{document}